\documentclass[11pt]{article}
\usepackage{geometry}
\usepackage{coling2020}
\usepackage{times}
\usepackage{url}
\usepackage{latexsym}
\usepackage{microtype}

\usepackage{graphicx}
\usepackage{subcaption}
\usepackage{tabularx}
\usepackage{array}
\usepackage{multirow}
\usepackage{color}
\usepackage{xcolor}
\usepackage{soul}

\usepackage{cleveref}
\crefname{section}{§}{§§}
\Crefname{section}{§}{§§}

\setstcolor{red}

\colingfinalcopy 

\title{newsSweeper at SemEval-2020 Task 11: Context-Aware Rich Feature Representations For Propaganda Classification}

\author{Paramansh Singh$^{*}$ \qquad    
  Siraj Sandhu$^{*}$  \qquad  
  Subham Kumar\thanks{\quad Authors equally contributed  to this work.} \qquad
  Ashutosh Modi  \\
{Indian Institute of Technology Kanpur (IITK)} \\
  {\tt \{params,sssandhu,subhamkr\}@iitk.ac.in}  \\
  {\tt ashutoshm@cse.iitk.ac.in}  \\
}

\date{}

\begin{document}
\maketitle

\begin{abstract}
  This paper describes our submissions to SemEval 2020 Task 11: \textit{Detection of Propaganda Techniques in News Articles} for each of the two subtasks of \textit{Span Identification} and \textit{Technique Classification}. We make use of pre-trained BERT language model enhanced with tagging techniques developed for the task of Named Entity Recognition (NER), to develop a system for identifying propaganda spans in the text.
  For the second subtask, we incorporate contextual features in a pre-trained RoBERTa model for the classification of propaganda techniques.
We were ranked 5$^{th}$ in the propaganda technique classification subtask.

\end{abstract}

\section{Introduction}
\label{intro}

%
%
\blfootnote{
    \hspace{-0.65cm}  
    This work is licensed under a Creative Commons 
    Attribution 4.0 International License.
    License details:
    \url{http://creativecommons.org/licenses/by/4.0/}.
}

Propaganda is commonly defined as information of a biased or misleading nature, possibly purposefully shaped, to promote an agenda or a cause.
Commonly used propaganda techniques are psychological and rhetorical - ranging from the selective presentation of facts and logical fallacies to the use of loaded language to produce an emotional response. Historically, propaganda has been widely employed and is often associated with governments, activists, big business, religious organizations, and partisan media.

\begin{figure}[h]
    \includegraphics[width=\textwidth]{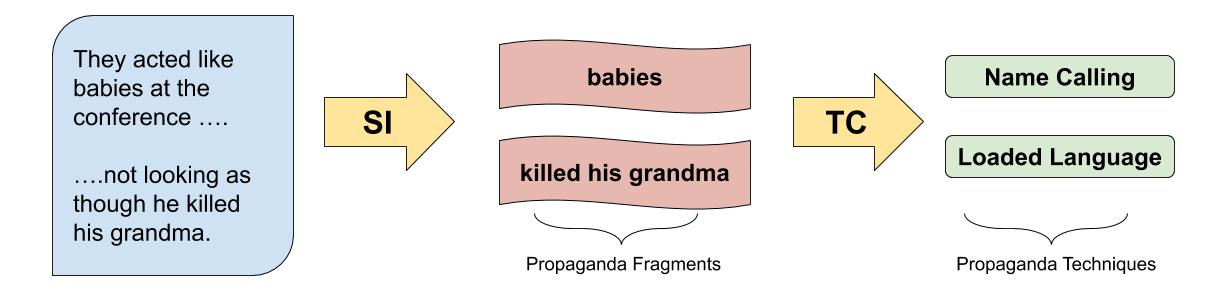}
    \caption{Example of subtasks \textit{span identification}(SI) and \textit{technique classification}(TC)}
    \label{fig:subtasks_eg}
\end{figure}

Having been first introduced in a previous iteration of the SemEval shared task \cite{martino2019findings}, the current shared task \cite{DaSanMartinoSemeval20task11} consists of two subtasks, \textbf{(i) Span Identification(SI):} Given a plain-text document, identify those specific fragments which are propagandistic,
and \textbf{(ii) Technique Classification(TC):} Given a text fragment identified as propaganda and its document context, identify the applied propaganda technique in the fragment. Both the subtasks focus on English texts only. An illustration is shown in Figure \ref{fig:subtasks_eg}.

The previous iteration of this task required identifying the propaganda spans and also identifying the propaganda technique for that fragment.  
The highest scoring system by \newcite{yoosuf-yang-2019-fine} employed an out-of-the-box BERT \cite{DBLP:journals/corr/abs-1810-04805} language model fine-tuned on the token classification task - whether the token is part of a propaganda span or not. For the classification task, the same system employed an N+2 way classification, where N is the number of propaganda classes. For the technique classification, we make use of document context. For context inclusion, we build upon an approach by \newcite{hou-chen-2019-caunlp}, where they use title and previous sentence as context. They pass the context concatenated with the original text for fine-tuning the BERT model.

Our approach for span identification makes use of a state-of-the-art language model enhanced by tagging schemes inspired from Named Entity Recognition which aim to model spans better - logically and intuitively - by involving \textit{Begin} and \textit{End} tags \cite{ramshaw-marcus-1995-text} to better formulate spans. In this regard, we experimented with BERT, RoBERTa \cite{DBLP:journals/corr/abs-1907-11692}, SpanBERT \cite{DBLP:journals/corr/abs-1907-10529}, GPT2 \cite{Radford2019LanguageMA} language models and with  BIO, BIOE, BIOES tagging schemes. For the final technique classification model, we use RoBERTa language model to get the contextual sequence representation for the propaganda fragment and perform classification.

Our best performing models ranked 13$^{th}$ for SI subtask and 5$^{th}$ for the TC subtask on an unknown test set. The implementation for our system is made available via Github\footnote{\url{https://github.com/paramansh/propaganda_detection}}.


\section{Problem Statement}
For the SI subtask, given a document $\mathbf{D}$ we have to identify all such fragments of text which are propagandistic.
These character level input spans are converted to token level labels (see \S \ref{sec:si_preprocess}), which makes this problem a token classification problem. The approach has been detailed out in \cref{sec:si_sys_des}. For the TC subtask, given a span identified as propagandistic, we have to classify it into one of the 14 techniques \cite{DaSanMartinoSemeval20task11}. Note that we are also given the document $\mathbf{D}$ containing the span which we can use as context. The approach has been detailed out in \cref{sec:tc_sys_des}. In \cref{sec:dataset} we describe the dataset.

\section{System Description}
\subsection{Span Identification}
\label{sec:si_sys_des}
Span Identification is a binary sequence tagging task where we classify each token in a sequence as a propaganda/non-propaganda text (P/NP), that is whether it is part of a propaganda fragment. For this, we pass the input sequence to a pre-trained Transformer based model (for example, BERT, RoBERTa etc.) and get embeddings for each token in the sequence. 
These embeddings are then passed to the classification layer on top of the Transformer which classifies the token as P/NP. 
The overall architecture of the system is shown in Figure \ref{fig:subim1}. Note that the classifier weights shown are shared across all tokens. 

\subsubsection*{Tagging}
\begin{figure}[h]
\begin{subfigure}{0.5\textwidth}
\includegraphics[width=0.9\linewidth]{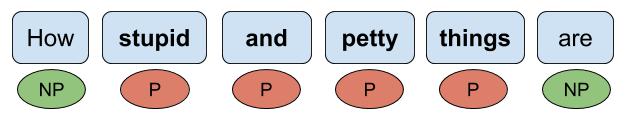} 
\caption{P/NP tagging}
\label{fig:PNP}
\end{subfigure}
\begin{subfigure}{0.5\textwidth}
\includegraphics[width=0.9\linewidth]{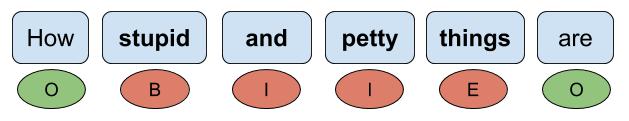} 
\caption{BIOE tagging scheme}
\label{fig:BIOE}
\end{subfigure}
\caption{Comparison of the two tagging schemes. The text shown in bold is a propaganda fragment.}
\label{fig:SI}
\end{figure}

In the approach described above, each token in the sequence is tagged as P/NP. 
We take this approach forward and tag using BIOE scheme. 
In this scheme, \textbf{B} represents the \textit{Beginning} of propaganda text, \textbf{I} means \textit{Inside} propaganda text, \textbf{E} represents the \textit{End} of propaganda text,  and \textbf{O} means \textit{outside} the propaganda text i.e., non-propaganda text. 
The comparison between the texts tagged using the two schemes is shown in Figure \ref{fig:SI}.
For span identification, the rest of the model architecture is the same, but the tokens are now labelled as BIOE instead of just P/NP earlier.
During prediction, the tokens labeled as \textbf{B}, \textbf{I}, or \textbf{E} are classified as propaganda token. 

Begin and End tags help to formulate the notion of spans better and model it as a \textit{span identification problem} rather than a \textit{token level classification problem} by introducing dependencies between the various tokens part of a propaganda span. 

\subsection{Technique Classification}
\label{sec:tc_sys_des}
Technique classification is a multiclass sequence classification task, where we identify the propaganda technique for a given propaganda fragment. In addition to the fragment, we also have the document containing the fragment. To solve this task, we first pass the propaganda sequence and its context to a pre-trained Transformer. We use the Transformer outputs to get the sequence representation $\mathbf{S}$ and  context representation  $\mathbf{C}$.  Both of these are combined to get the contextualized vector representation $\mathbf{V}$ of the sequence. This vector representation $\mathbf{V}$ is then passed to the classifier layer on top, which performs the final classification (refer Figure \ref{fig:subim3}). 

\subsubsection*{Sequence and Context Vectors} We use pre-trained RoBERTa to obtain the representation for the sequence via the CLS output vector. The CLS output vector gives the aggregate sequence representation as modeled in \newcite{DBLP:journals/corr/abs-1810-04805}.

The context surrounding the span can be at different granularity: article level, paragraph level, or sentence level. Obtaining vector representation for a  paragraph or an article context using BERT or other language models is difficult as articles can be very long. As a result, the article context vector is obtained using the CLS representation of the headline of the news article as suggested by \newcite{hou-chen-2019-caunlp}.
To capture the sentence context, we use the sentence corresponding to the propaganda fragment. If the fragment spans across multiple sentences, all such sentences are considered. However, we limit the length of sentence contexts to 130 words. This is because, for sufficiently long propaganda spans, the exact meaning can be inferred directly from the text without any surrounding context.

\subsubsection*{Contextual Sequence Representation}
\begin{figure}[h]

\begin{subfigure}{0.49\textwidth}
\includegraphics[width=0.9\linewidth, height=4cm]{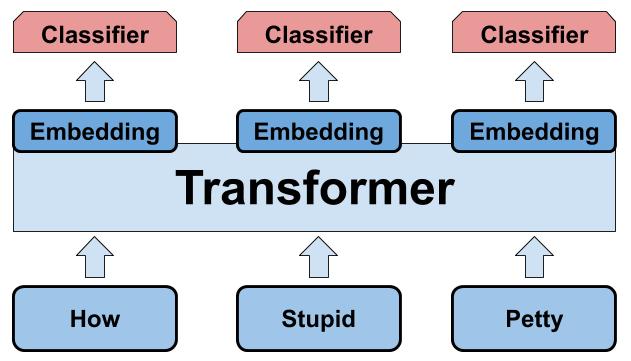} 
\caption{Span Identification}
\label{fig:subim1}
\end{subfigure}
\begin{subfigure}{0.55\textwidth}
\includegraphics[width=0.9\linewidth, height=4cm]{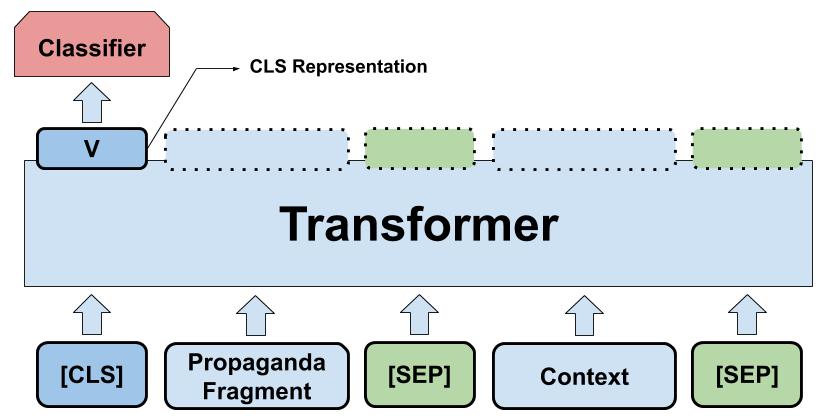}
\caption{Technique Classification - System 1}
\label{fig:subim2}
\end{subfigure}
\begin{subfigure}{\textwidth}
\includegraphics[width=\linewidth]{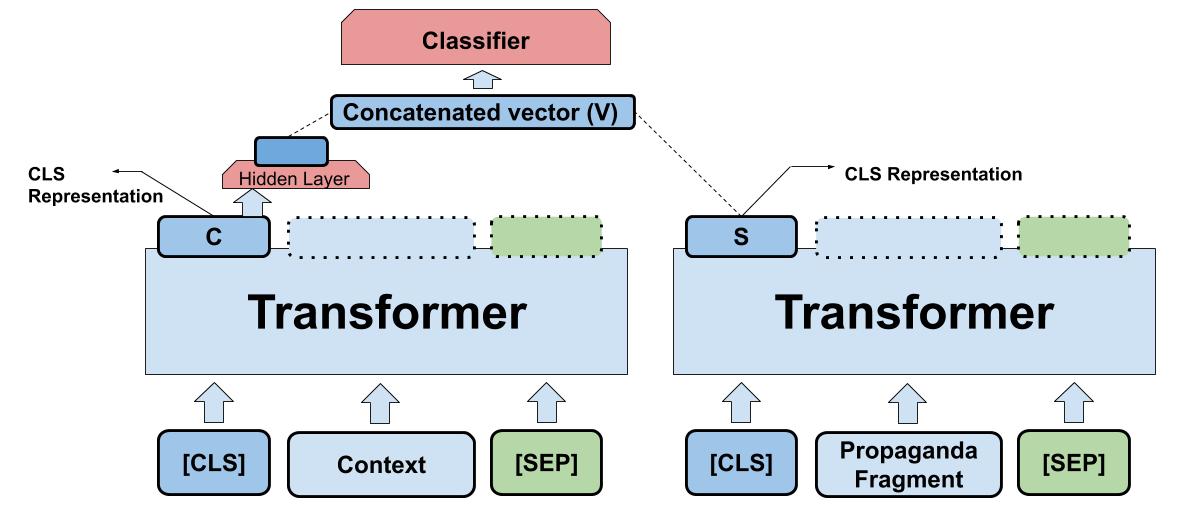} 
\caption{Technique Classification - System 2}
\label{fig:subim3}
\end{subfigure}
\caption{Systems for SI and TC tasks. Note that Transformer can be any pre-trained language model like BERT, RoBERTa, SpanBERT etc.}
\label{fig:image21}
\end{figure}

We followed two approaches to include context. In the first system, we pass the context text concatenated with the propaganda fragment text directly to the Transformer (refer Figure \ref{fig:subim2}). We take the resultant sequence embedding output from the Transformer as the contextual representation ($\mathbf{V}$) of the fragment. In this case, $\mathbf{S}$ and $\mathbf{C}$ are not explicitly generated. One problem with this approach is that in case of small propaganda fragments (2 or 3 words), the longer context text will influence the final representation, which may not be ideal.

We address this in System 2, where we pass the fragment and the context to different Transformers to get $\mathbf{S}$ and  $\mathbf{C}$ independently. There is an additional hidden layer on top of the context Transformer which reduces the dimension of context vector $\mathbf{C}$. This resultant vector is then concatenated with $\mathbf{S}$ to get  $\mathbf{V}$. The additional hidden layer allows the classifier to give more attention to the propaganda sequence. This system is shown in Figure \ref{fig:subim3}.
Apart from concatenation, we tried another approach where we set $\mathbf{V} = \alpha\mathbf{S} + (1-\alpha)\mathbf{C}$, i.e., a weighted average of $\mathbf{S}$ and $\mathbf{C}$.

\section{Experimental Setup}
\subsection{Data}
\label{sec:dataset}
The dataset is released by the Semeval task organizers \cite{EMNLP19DaSanMartino}. The input for both tasks will be news articles in plain text format (unimodal).
The gold labels for SI are in the form of an article identifier, a begin and an end character offset for each propaganda instance. The labels for TC are in the same format, but with an additional field for the class annotation pertaining to each instance. Note that if a text fragment falls into belongs to multiple classes, the same occurs as many times in the corpus with a single class label in each instance.
\subsection{Pre-Processing}
\subsubsection{Span Identification}
\label{sec:si_preprocess}
The articles are too long to directly feed standard language models and are thus split by sentences. This resulted in 17,855 sequences with a maximum length of 179 words. 
We convert the character level span labels to token level labels where a token labeled as positive implies that the token is part of a propaganda span. As a result, our training objective becomes classifying each token as a propaganda word or not. In the training data, approximately 13 percent of tokens were part of a propaganda text. During prediction, the consecutive tokens with the same label are merged to get the final character level span outputs. 

\subsubsection{Technique Classification}
For this task, we get the propaganda fragment text using the given character labels. For getting the surrounding sentence context, we limit the maximum number of words to 130. We select words from both sides in our context until the end of the sentence is reached or the maximum word limit is reached.
The training data consists of (sentence, context) pair and the propaganda technique is the label.
We get 6129 such pairs in our training dataset. The dataset is highly unbalanced w.r.t the labels, for example, the category \texttt{Loaded Language} has 2123 instances, and \texttt{Thought-terminating\_Cliche} has only 76 instances. 

\subsection{Training Details}
We use PyTorch framework provided by huggingface\footnote{\url{https://huggingface.co}} library for the pre-trained BERT and RoBERTa models. We use \texttt{bert-base-uncased} for span identification task, and \texttt{roberta-uncased} for Technique Classification task. In addition to this, we experiment with GPT-2 and SpanBERT as well. For both tasks, we fine-tuned using AdamW optimizer \cite{DBLP:journals/corr/abs-1711-05101} for 4 epochs, with a learning rate of $3 \times 10^{-5}$ and batch size = 8.

\section{Results}
\subsection{Span Identification}
For this task we experimented with the following, (i) Tagging Scheme:  P/NP, BIO, BIOE and BIOES, (ii) Language Models: BERT (base uncased, large uncased), RoBERTa, GPT2, and SpanBERT, (iii) Concatenating different hidden layers of BERT to get token representation, (iv) Inclusion of linguistic features using POS tags of tokens. The evaluation metric used to compare is F1 score, which is calculated based on the overlap between the predicted and actual spans. The scores for different models on official development set trained with 90 percent training data have been reported in Table \ref{tab:SI-results}.

\newcolumntype{P}[1]{>{\centering\arraybackslash}p{#1}}

\begin{table}[]
{\renewcommand{\arraystretch}{1}
\begin{tabular}{|p{0.30\textwidth}|P{0.18\textwidth}|P{0.2\textwidth}|P{0.2\textwidth}|}

\hline
\textbf{Model}                                                           & \textbf{F1 Score}            & \textbf{Precision}           & \textbf{Recall}              \\ \hline
BERT - P/NP                                                              & 0.414                      & 0.401                      & 0.427                      \\ \hline
BERT - BIO                                                               & 0.434                      & 0.403                      & 0.471                      \\ \hline
BERT - BIOE                                                              & \textbf{0.445}             & 0.387                      & 0.521                      \\ \hline
BERT - BIOES                                                             & 0.411                      & 0.431                      & 0.392                      \\ \hline
BERT\_large - BIOE                                                      & 0.411                      & 0.416                      & 0.407                      \\ \hline
GPT2 - BIO                                                               & 0.256                      & 0.298                      & 0.225                      \\ \hline
SpanBERT - BIOE & 0.445 &	0.398	& 0.504  \\ \hline
RoBERTa - BIOE & 0.429 & 0.403	& 0.458 \\ \hline
BERT - BIOE - POS tags                                                               & 0.411                      & 0.431                      & 0.392 \\ \hline
BERT - BIOE - 4 layer concat & 0.424 & 0.367 & 0.502 \\ \hline
BERT - BIOE - 12 layer concat & 0.399 & 0.436 & 0.368 \\ \hline
\end{tabular}
}
\caption{Task SI Results on development set trained with 90 percent of training data.}
\label{tab:SI-results}
\end{table}

As it can be seen, the score improved on changing the tagging scheme from P/NP to BIO and from BIO to BIOE. As suggested earlier, this can be because, BIO, BIOE capture the span nature of the output. One way to realize this is by analyzing predictions on development set, where the average output length of spans using P/NP tagging is 33 chars, for BIOE it is 36 chars, compared to the actual average length which is 38 chars.

However, none of the other approaches resulted in any improvement. Due to less number of examples in the training data, all the models were overfitting and failed to generalize. We used the standard BERT-BIOE system for our final test set submission.

\subsection{Technique Classification}
In this we experiment with the following, (i) Language Models: BERT and RoBERTa, (ii) two approaches to include context as described in \S~\ref{sec:tc_sys_des}, (iii) context to include: sentence level or document level (from article headline), and (iv) methods to combine the embeddings in System 2: default method, without hidden layer concatenation, and weighted average. The evaluation metric used to compare is micro-averaged F1 score. The results are reported for the official development set in Table \ref{tab:tc-results}. The Baseline model is the one not using contextual features.
\begin{table}[]
{\renewcommand{\arraystretch}{1}}
\resizebox{\textwidth}{!}{%
\begin{tabular}{|l|l|l|l|}
\hline
\textbf{Model}                    & \textbf{F1} & \textbf{Model}                    & \textbf{F1} \\ \hline
BERT Baseline                     & 0.583             & RoBERTa Baseline                  & 0.602             \\ \hline
RoBERTa - Title - System 1 &  0.578     & RoBERTa - Sentence - System 1 & 0.593  \\ \hline
RoBERTa - Title - System 2 w/o Hidden & 0.601 & RoBERTa - Sentence - System 2 w/o Hidden &  0.599 \\ \hline
RoBERTa - Sentence - System 2 - Add & 0.598 & RoBERTa - Sentence - System 2 &  \textbf{0.611} \\ \hline
RoBERTa - Baseline - Length Feature & 0.589 & RoBERTa - Sentence - System 2 - Weighted Avg.  & 0.593  \\ \hline
\end{tabular}%
}
\caption{Task TC Results on development set trained with 90 percent of training data. Baseline refers to model without contextual features. System 1 (Figure \ref{fig:subim2}) - including context by concatenating text. System 2 (Figure \ref{fig:subim3}) - including context by concatenating embeddings, with and without hidden layer above context vector.}
\label{tab:tc-results}
\end{table}

We observed that for this task RoBERTa outperforms BERT and was thus used to generate input representations. Also, context inclusion did not result in any significant improvements in most cases. System 1 using sentence as context lead to a significant drop in F1 score, which can be attributed to the fact that the longer context sentence takes the attention away from the actual text to be classified. Other than that, most other methods performed similar to the baseline. However, System 2 with a hidden layer on top of the context vector, resulted in slightly better performance as it put more emphasis on the sequence to be classified compared to the surrounding context, which in some cases may lead to poor performance as mentioned above. 
Although we can expect the model to learn this after directly concatenating as well, it may not due to the lack of sufficient data. 
By slightly tuning hyperparameters and using all of the data for the RoBERTa baseline model, we were able to achieve an F1 score of 0.627 on the development set.
We used this fine-tuned system for our test set submission.

\subsubsection*{Data Imbalance}
All the models displayed a significant difference between the class-wise maximum and minimum F1 scores. Techniques such as \texttt{Loaded\_Language} and \texttt{Name\_Calling,Labeling} with many training examples had F1 scores close to 0.7. On the other hand, minority classes, especially the ones which are a combination of multiple techniques (\texttt{Whataboutism,Straw\_Men,Red\_Herrin} and \texttt{Bandwagon,Reductio\_ad\_hitlerum}), had F1 scores less than 0.1.

\subsubsection*{Span Length Distribution}
\begin{figure}[h]

\begin{subfigure}{0.33\textwidth}
\includegraphics[width=\linewidth]{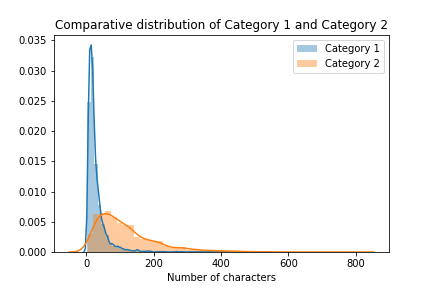} 
\caption{Category 1 vs Category 2}
\label{fig:comparative}
\end{subfigure}
\begin{subfigure}{0.33\textwidth}
\includegraphics[width=\linewidth]{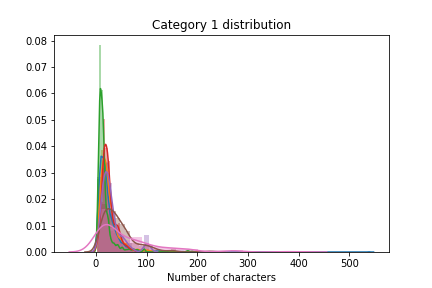} 
\caption{Category 1 techniques}
\label{fig:category1}
\end{subfigure}
\begin{subfigure}{0.33\textwidth}
\includegraphics[width=\linewidth]{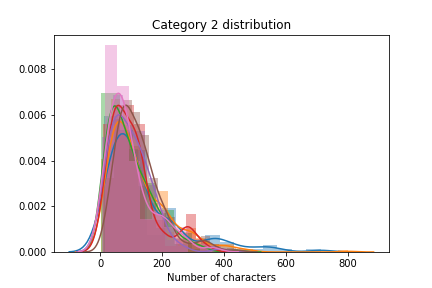} 
\caption{Category 2 techniques}
\label{fig:category2}
\end{subfigure}
\caption{Distribution of span length for various categories}
\label{fig:lengths}
\end{figure}

Figure \ref{fig:lengths} details the distribution of lengths of spans for various categories. Category 1 \footnote{Category 1 - Loaded\_Language; Name\_Calling,Labeling; Repetition; Slogans; Thought-terminating\_Cliches; Exaggeration,Minimisation; Flag-Waving} consists of 7 classes, while the remaining 7 are part of category 2. As the figure suggests, Category 1 techniques follow similar, more ``peaky" distribution, while Category 2 techniques follow a similar ``flat" distribution. We tried to model this by passing length as a feature after at the final classification layer and trying separate models for both, but none could beat our baseline model. 

\section{Conclusion}
We proposed models for detecting propaganda fragments in articles and classifying the propaganda technique used in a given propaganda fragment. We portray how state-of-the-art language models can be useful for both subtasks. We show how BIOE tagging scheme can help detect spans better.
For classification, we model a way to include context surrounding a propaganda fragment, although its inclusion does not improve the predictions significantly. Finally, we also show how the span lengths are distributed for different categories. Modeling this fact by doing hierarchical classification is something that can be explored in the future. 
\newpage
\bibliographystyle{coling}
\bibliography{semeval2020}


\end{document}